\documentclass{article}
\usepackage{spconf,amsmath,amssymb,amsfonts,graphicx}

\usepackage{epsfig} 
\usepackage{times} 
\usepackage{commath}
\usepackage{bbm}
\usepackage{dsfont}
\usepackage{placeins}

\usepackage{color,colortbl}
\usepackage[table]{xcolor}

\usepackage{multirow}
\usepackage{makecell}
\usepackage{booktabs}
\usepackage{svg}

\usepackage{cite}
\usepackage{url}
\usepackage[hidelinks]{hyperref}
\hypersetup{
    colorlinks=true,
    linkcolor=black,
    filecolor=black,
    urlcolor=gray,
    citecolor=black,
    breaklinks=true,
}

\usepackage{tikz}
\usepackage{pgfplots}
\pgfplotsset{width=10cm,compat=1.15}
\usepgfplotslibrary{statistics}
\usetikzlibrary{pgfplots.groupplots}

\definecolor{mylightgray}{gray}{0.9}

\definecolor{c9}{RGB}{0 158 115} 

\definecolor{mybrown}{RGB}{165 80 80} 
\definecolor{c1}{RGB}{0,119,187}  
\definecolor{c2}{RGB}{51,187,238} 
\definecolor{teal}{RGB}{0,238,55}  
\definecolor{c4}{RGB}{238,119,51} 
\definecolor{c5}{RGB}{204,51,17} 
\definecolor{c6}{RGB}{238,51,119} 
\definecolor{c7}{RGB}{187,187,187} 
\definecolor{c8}{RGB}{0 0 0} 
\definecolor{mygreen}{RGB}{50,250,50} 

\definecolor{myblack}{RGB}{0 0 0} 
\definecolor{mygrey}{RGB}{128 128 128} 

\definecolor{red1}{RGB}{240,128,128}
\definecolor{red2}{RGB}{255,69,0}
\definecolor{red3}{RGB}{139,0,139}

\title{A mixed-reality dataset for category-level 6D pose and size estimation of hand-occluded containers}
\name{Xavier Weber, Alessio Xompero, Andrea Cavallaro\thanks{This work is supported by the CHIST-ERA program through the project CORSMAL, under UK EPSRC grant EP/S031715/1.}}
\address{Centre for Intelligent Sensing, Queen Mary University of London, UK}
\begin{document}
\ninept
\maketitle
\begin{abstract}
Estimating the 6D pose and size of household containers is challenging due to large intra-class variations in the object properties, such as shape, size, appearance, and transparency. The task is made more difficult when these objects are held and manipulated by a person due to varying degrees of hand occlusions caused by the type of grasps and by the viewpoint of the camera observing the person holding the object. In this paper, we present a mixed-reality dataset of hand-occluded containers for category-level 6D object pose and size estimation. The dataset consists of 138,240 images of rendered hands and forearms holding 48 synthetic objects, split into 3 grasp categories over 30 real backgrounds. We re-train and test an existing model for 6D object pose estimation on our mixed-reality dataset. We discuss the impact of the use of this dataset in improving the task of 6D pose and size estimation.
\end{abstract}
%
\section{Introduction}
\label{sec:intro}

Estimating the position and orientation in 3D (or 6D pose) of an object from a single view is important for augmented reality, immersive online gaming, online shopping, robotic grasping and manipulation, or human-to-robot handovers in assistive scenarios~\cite{Sanchez-Matilla2020,Ortenzi2021TRO,tremblay2018deep,Wang2019CVPR_NOCS}. Despite being investigated for a long time, the problem is still challenging and far from solved due to the large variety of objects and their physical properties, the differences in environmental conditions, and the uncertainty in the applicative scenarios. For example, objects vary in their shape, size, appearance (different textures or lack of texture, colour), and transparency~\cite{hodan2018bop,labbe2020cosypose,Xiang2018RSS_PoseCNN}. These physical properties can also vary within instances of a same category, such as food boxes, containers for liquids, cameras, or shoes~\cite{Wang2019CVPR_NOCS,objectron2020}. Moreover, some objects can be manipulated by people (e.g., in the kitchen~\cite{Damen2022RESCALING}), resulting in increased and more uncertain variations in the object poses and different degrees of hand occlusions.

\begin{figure}[t!]
    \centering
    \includegraphics[width=0.9\columnwidth]
    {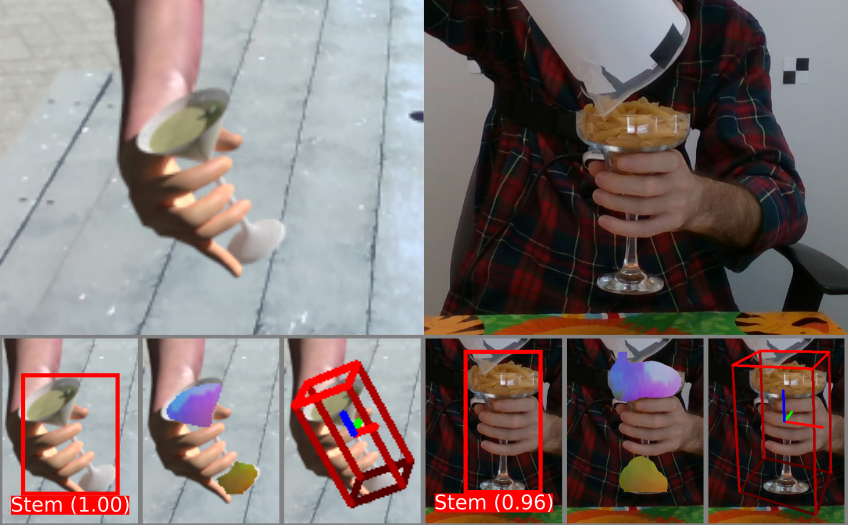}
    \caption{6D pose estimation for a container manipulated by a person on our mixed-reality (left) and on CORSMAL Containers Manipulation~\cite{Xompero2022Access} (right) datasets. A model~\cite{Wang2019CVPR_NOCS} can first identify and localise the container on the image, predict the 3D normalised coordinates, and then recover the pose and size of the object in 3D.}
    \label{fig:visres}
    \vspace{-10pt}
\end{figure}

Early approaches rely on the availability of 3D models of opaque objects with enough texture to identify corner-like points in the image and estimate the pose with a close-form solution or an optimisation strategy~\cite{Lepetit2009IJCV}. Approaches based on deep neural networks use large and diverse amounts of data to train a model to predict the pose of objects with different properties~\cite{hodan2018bop,labbe2020cosypose,Liu2020CVPR_KeyPose}. The task is nowadays often combined with segmentation of the object(s) in the image, or with other tasks, such as shape reconstruction~\cite{Wang2019CVPR_NOCS,labbe2020cosypose,irshad2022shapo} (see Fig.~\ref{fig:visres}). 
%
Because of the dependency on data to train the models, there has been an increase of publicly available datasets for object pose estimation. Each dataset focuses on a specific property of a set of objects or object categories (e.g., lack of texture, shape variations, transparency), or on a specific purpose (e.g., grasping objects lying on a surface, reconstruction of objects and hands during a human manipulation of the object). Examples of these datasets are LineMOD~\cite{hinterstoisser2012model}, HomebrewedDB~\cite{kaskman2019homebreweddb}, YCB-Video~\cite{Xiang2018RSS_PoseCNN}, TUD-L~\cite{hodan2018bop} TOD~\cite{Liu2020CVPR_KeyPose}, HO3D~\cite{Hampali2020CVPR_HO3D}, NOCS~\cite{Wang2019CVPR_NOCS}, Objectron~\cite{objectron2020}, and OBMan~\cite{Hasson2019CPR_obman}. TOD and YCB-Video focus on specific object instances and therefore cannot be used to generalise to large variability between objects of the same class. Objectron and NOCS consist of image sequences, captured by a moving camera, of single or multiple objects lying on a surface for estimating the pose and size of these objects belonging to a set of categories. Unlike a cut-and-paste approach~\cite{Dwibedi_2017_ICCV} that cannot provide annotations of the 6D object pose, NOCS also generates images of synthetic objects rendered on real background scenes in a context-manner (mixed-reality) to handle the necessity of large data with corresponding free annotations of the 3D normalised coordinated maps. However, all the above-mentioned datasets do not include manipulation of the objects by a person, thus limiting the variability in the object poses to only those related to the camera motion. 
HO3D and OBMan address handheld objects for reconstructing or estimating the pose of both the human hand and the manipulated object. However, HO3D considers a limited subset of object instances from the list of YCB objects, therefore lacking any intra-class variability. OBMan is a category-level based and object-centric dataset that randomly renders synthetic objects and hands on top of real backgrounds. Datasets such as OBMan and the mixed-reality part of NOCS avoid the collection of real data and their annotation that may be cumbersome, span a large amount of time, and require the involvement of many people as well as the use of additional (expensive) equipment, while allowing the generation of large composite image data accompanied by free annotations. However, the random-based data augmentation approach to generate OBMan lacks a context-based composition with the scene and plausible and structured grasps suitable, for example, for a collaboration. 
Epic-Kitchens is another large audio-visual and text dataset with a large variety of objects that are manipulated by the person in the kitchen environment~\cite{Damen2022RESCALING}. However, the dataset focuses on action recognition from first person view and is not suitable for pose estimation as it lacks the corresponding annotations. CORSMAL Containers Manipulation (CCM) is a multi-view dataset consisting of videos of people manipulating various household containers, e.g., pouring a content into a cup or drinking glass, or shaking an already filled food box~\cite{Xompero2022Access}. However, the number of instances for each category is limited (15), and annotations of the object poses and corresponding 3D models are not available. 
Despite all of these efforts, there is still no large dataset for 6D pose estimation of objects that vary in their physical properties and with different degrees of hand occlusions based on how a person would grasp these objects. 

\begin{figure}[t!]
    \centering
    \includegraphics[width=\columnwidth]{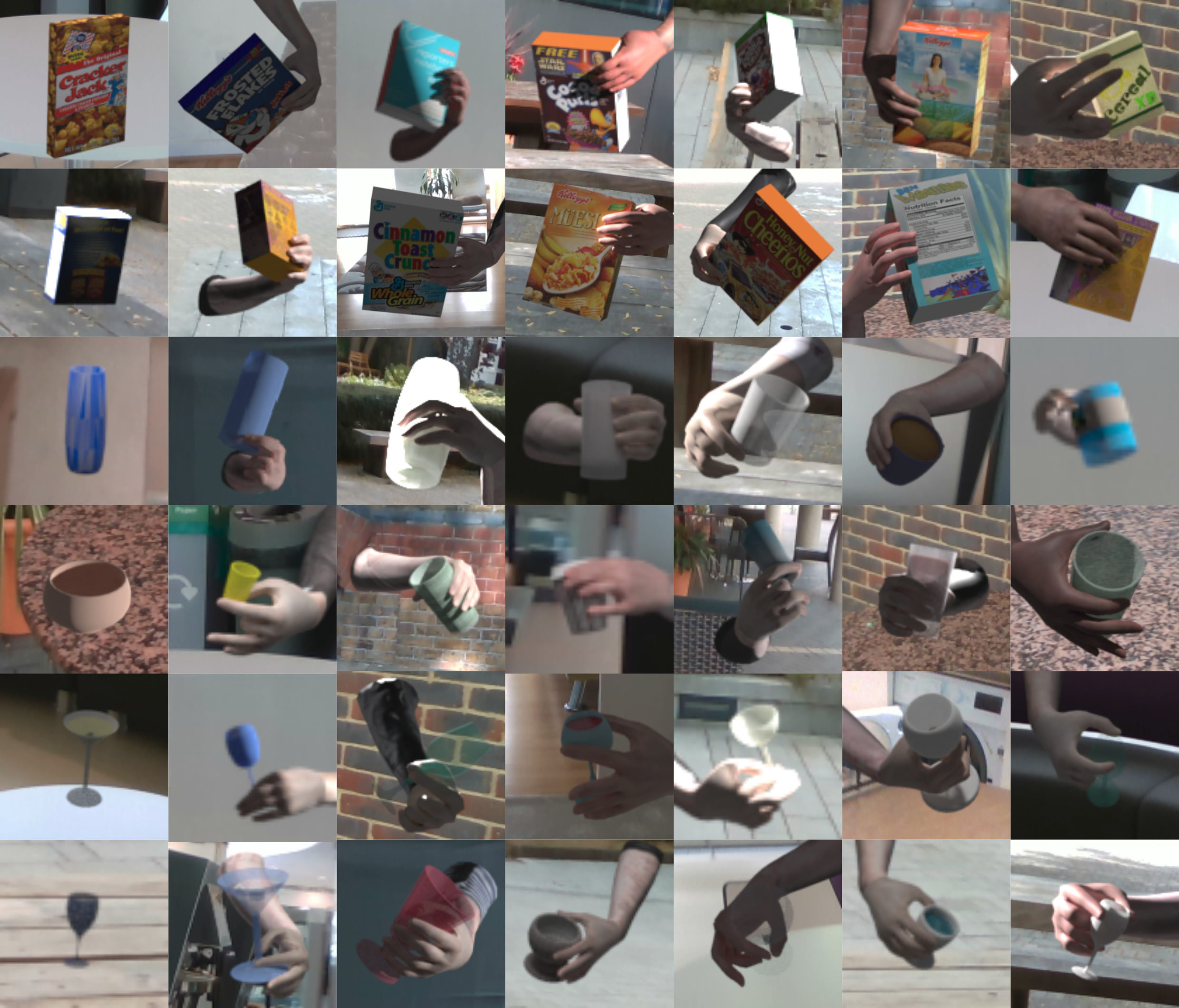}
    \caption{Sample of image crops from the generated mixed-reality dataset of containers and hand-held containers. Note the diversity in human grasps, colour and illumination on the rendered hand and forearms, and in object texture and transparency. }
    \label{fig:datasetsamples}
    \vspace{-10pt}
\end{figure}

In this paper, we present a new mixed-reality dataset for category-level pose and size estimation of containers belonging to a set of categories (food boxes, cups and drinking glasses) and held by a human hand. Fig.~\ref{fig:datasetsamples} shows examples of image crops of the dataset. The dataset exhibits varying degrees of occlusions and well-defined plausible grasp types (top, bottom, natural) with the purpose of a collaboration~\cite{Sanchez-Matilla2020}. Specifically, we generate the mixed-reality dataset in a pseudo-realistic manner by rendering synthetic handheld objects on top of real backgrounds, while accounting for their location in the scene, and with plausible illumination, resembling the lights in the background scene. 
We also provide an annotation of 60 sequences for each of the three fixed views from CCM and show how our mixed-reality dataset enables the training and deployment of deep learning models in real settings. To this end, we validate an existing approach (i.e., NOCS~\cite{Wang2019CVPR_NOCS}) on a test split of the mixed-reality dataset and on the annotated real sequences\footnote{Data, models, code, and additional results will be available at\\ \url{https://corsmal.eecs.qmul.ac.uk/pose.html}}. 




\section{Mixed-reality dataset}

In this section, we present our procedure to generate pseudo-realistic composite images of handheld containers to form the mixed-reality dataset for category-level object pose and size estimation in 3D.  

We select containers that can be held or manipulated by a person, such as food boxes and drinking glasses or cups, and of which the physical properties highly vary in size, shape, transparency and textures (or absence of texture)~\cite{Xompero2022Access}. However, cups and drinking glasses differ only in their material, making the recognition of these two types of objects hard for a vision model. Therefore, we categorise the cups and drinking glasses based on their geometry, and in particular by the presence of a stem. 
The categories are \textit{human}, \textit{non-stem}, \textit{stem}, \textit{box}, and we include \textit{background} for anything else~\cite{He2017ICCV_MaskRCNN}.

As \textit{objects}, we select 16 CAD models\footnote{Maximum number of unique \textit{stem}-like objects in ShapeNetSem.} from ShapeNetSem with their corresponding textures for each object category~\cite{savva2015semgeo}. We scale the objects to realistic dimensions by fixing the aspect ratio and sampling the height in the interval $[10,32]$~cm for \textit{box}, $[5,18]$~cm for \textit{non-stem}, and $[6,20]$~cm for \textit{stem}. 

As \textit{grasp types}, we follow the CORSMAL benchmark~\cite{Sanchez-Matilla2020} and we define six ways to hold an object based on the used hand and the position of the hand on the object: grasp at the bottom with left hand; grasp on top with left hand; natural grasp with left hand; grasp at the bottom with right hand; grasp on top with right hand; and, natural grasp with right hand. Specifically, we use the MANO hand model~\cite{romero2017embodied} and the GraspIt!~\cite{miller2004graspit} tool to manually generate right-hand grasps for each of the 48 objects, and we mirror the right-hand grasps to generate the left-hand versions. Because of the varying object shapes and sizes, we manually annotate a total of 288 grasps.


As \textit{backgrounds}, we acquire 30 images with an Intel D435i RealSense sensor in 10 different scenes (5 outdoor, 5 indoor) under 3 different views. Each background contains a flat surface -- e.g., a table or counter -- where the (handheld) objects are rendered. The sensor provides spatially aligned RGB and depth images with a resolution of 640x480 pixels. Depth images are captured up to a maximum distance of 6~m, and we apply spatial smoothing, temporal smoothing, hole filling, and decimation as filters during acquisition. 
For each background, we manually annotate a lighting setup using Blender~\cite{blender} to resemble the real scene conditions as close as we can when rendering the object on top of the backgrounds. We use sun-like source light for outdoor scenarios, whereas the lighting setup can include multiple omnidirectional and directional point-like source lights and rectangle area-base lights for indoor scenarios to reproduce bulb-lights, LEDs, and windows. These source lights are adjusted in terms of position, orientation, colour and energy strength. 

We generate a total of 138,240 composite images, split into 8,640 images of only objects on top of flat surfaces and 129,600 image of handheld objects above the flat surfaces. The dataset has 2,700 and 180 images for each object with grasps and without grasp, respectively. Images are evenly distributed for each grasp type (21,600), with 450 images for each object and grasp combination. 

For pseudo-realistically placing objects on top of flat surfaces in each background scene, we manually segment the flat surface on the image plane~\cite{labelme2016} and obtain the corresponding point cloud in 3D using the aligned depth map. We then compute the normal of the surface by automatically segmenting the 3D plane~\cite{silberman2012indoor} and removing outliers~\cite{Fischler1981ACM}. After randomly sampling from a uniform distribution a location on the flat surface where to place the object, we align the vertical axis of the object with the normal of the flat surface in the camera coordinate system and rotate the object around the vertical axis with a rotation randomly sampled from a uniform distribution in the range [0,359]$^\circ$. For each object and background combination, we sample 6 pairs of locations and orientations and render the objects on top of the background using Blender.

For pseudo-realistically rendering handheld objects, we place each object with a MANO hand and a forearm from the SMPL+H model~\cite{romero2017embodied}, based on the annotated grasp type, at randomly sampled locations up to 40~cm above the segmented flat surface for each background. We then rotate both the object and the hand by an amount randomly sampled from a uniform distribution in the interval $[-45,45]^\circ$ for pitch and roll, respectively. Note that for \textit{box}, we rotate the object of 180$^\circ$ around the vertical axis with a 50\% chance to render each side. 
To simulate a hand holding the object towards the camera, we place the forearm pointing towards the camera principal and rotate the forearm with an angle randomly sampled from a uniform distribution in the interval $[-45,45]^\circ$ around the vertical axis. For each object and grasp type, we sample 15 poses and render the handheld object on top of each background. Note that when rendering the forearms and hands, we randomly change the skin colour and the sleeves texture and type. We also avoid interpenetration between the composite mesh (object and the hand holding the object) and the flat surface by re-sampling another pose if any of the points belonging to the flat surface has a Euclidean distance smaller than 0.5 mm with respect to the nearest point on the composite mesh.

For each composite RGB image, we render accurate depth maps, segmentation masks for the containers and hands, and maps of the normalised coordinates for the containers~\cite{Wang2019CVPR_NOCS}, and we automatically annotate the 6D pose of the object.

\section{Validation}

In this section, we assess the benefit of our proposed mixed-reality dataset for category-level 6D object pose estimation. 
To this end, we modify the multi-branch deep learning model NOCS~\cite{Wang2019CVPR_NOCS} to recognise our categories. We compare the model that takes an RGB-D image as input with a modified post-processing stage to estimate the 6D pose and size of the containers from an RGB image with a prior.



\subsection{Experimental setup}

The original implementation of NOCS uses a ResNet-50~\cite{He2016CVPR} as a backbone, as there was no need to segment humans in their experimental settings. However, features about the human category are learnt better at deeper layer by using a ResNet-101 as in Mask R-CNN \cite{He2017ICCV_MaskRCNN}. To obtain an accurate segmentation of both the object and the person holding the object, we initialise the ResNet-101 backbone with pre-trained weights on COCO \cite{Lin2018ECCV_COCO}, and train the whole pipeline using images from our mixed-reality dataset; 1721 images of \textit{cup}, 1721 images of \textit{wine glass}, and 1721 images of \textit{human}, from COCO 2017; and 1,721 images of \textit{box} from Open Images v6~\cite{kuznetsova2020open}. Because of the different labelling, we map the COCO classes \textit{cup} and \textit{wine glass} to our categories \textit{non-stem} and \textit{stem}, respectively. Note that for Open Images we exclude images that contain people due to the lack of annotated person segmentation masks. To balance the data during training, we randomly sample 80\% of images from our dataset and 20\% of images from COCO and Open Images together. When training the 3D coordinates branch, the layers parameters are optimised only with the images and annotations from the mixed-reality dataset. 

Following the original training strategy~\cite{Wang2019CVPR_NOCS}, we train the model with their multi-objective function that uses a cross-entropy loss for classification~\cite{Girshick2015ICCV}, a smooth $\mathcal{L}_1$-norm loss for regressing the bounding box~\cite{Girshick2015ICCV}, an average binary cross-entropy loss for binary segmentation~\cite{He2017ICCV_MaskRCNN} and a symmetry-loss\footnote{Similar to NOCS~\cite{Wang2019CVPR_NOCS}, we select rotations with increments of 60$^\circ$ for \textit{non-stem} and \textit{stem}, and 180$^\circ$ for \textit{box} around the vertical axis.} for estimating the 3D normalised coordinates~\cite{Wang2019CVPR_NOCS}. For a single object, the total loss is
\begin{equation}
    \mathcal{L} = \mathcal{L}_{\text{cls}} + \lambda_1 \mathds{1} \mathcal{L}_{\text{bbox}} + \lambda_2 \mathcal{L}_{\text{seg}} + \lambda_3\mathcal{L}_{\text{sym}},
\end{equation}
where $\lambda_1,\lambda_2,\lambda_3$ are hyper-parameters that control the balance between the multiple losses~\cite{Girshick2015ICCV}, and the value of the indicator function $\mathds{1}$ is 0 when the category is \textit{background}.


NOCS uses Umeyama's algorithm~\cite{umeyama1991least} to estimate the pose and size of the object as the similarity transformation between the estimated 3D NOCS coordinates and the 3D coordinates obtained from the depth map of the segmented object. However, this assumes an accurate depth map that might not be available in the presence of sensor noise or transparent objects, i.e., missing or noisy depth values for the 3D point clouds can lead to wrong estimations of the object poses.
We observed that the predicted NOCS map intrinsically encodes the projection of the 3D normalised object coordinates into the image of a camera viewpoint and hence the pose of the object with respect to the canonical representation. We exploit this relationship to directly estimate the object pose up to a scale factor using EPnP~\cite{Lepetit2009IJCV} and the known camera intrinsic parameters. The dimensions of the object in normalised space are obtained by computing the absolute maximum after centring the object at the origin for each coordinated axis. To recover the scaling factor, we consider and compare three cases in addition to Umeyama's algorithm: using a naive prior that is the average scaling factor of a training object category selected by the NOCS classification branch (EPnP-A); the real true scaling factor of the object as a reference (EPnP-G); or the scaling factor estimated by Umeyama's algorithm (EPnP-U).

We use our mixed-reality dataset for training and testing the NOCS model.  We split the dataset into training, validation, and testing sets by leaving out six instances for validation and six instances for testing (two instances per category). This results in 103,680 images (36 instances) for training, 17,280  (6 instances) for validation, and 17,280 images  (6 instances) for testing.

For testing generalisation, we consider CCM~\cite{Xompero2022Access} and we select 60 sequences of a person manipulating a container as observed from three fixed views (two side and one frontal view). The sequences contain each of the fifteen containers under four randomly sampled diverse conditions, including background and lighting conditions, scenarios (person sitting, with the object on the table; person sitting and already holding the object; person standing while holding the container and then walking towards the table), and filling amount and type. Fifteen sequences exhibit the case of the empty container for all fifteen objects, whereas the other sequences have the person filling the container with either pasta, rice or water at 50\% or 90\% of the full container capacity. For each sequence, we manually annotate the 6D poses of the containers every 10 frames if visible in at least two views~\cite{Xompero2022ICASSP}. We evaluate the trained models with the four strategies on these annotated frames.

\subsection{Performance measures}
We evaluate the predicted poses and sizes using Average Precision:
\begin{equation}
    AP = \sum_{q=1}^Q (R_q - R_{q-1})\hat P(R_q),
\label{eq:ap}
\end{equation}
where $R_1,\ldots,R_Q$ are the recalls in ascending order (detected objects are sorted by 3D Intersection over Union (IoU) in descending order), and $Q$ is the number of detected objects. $\hat{P}$ is the interpolated precision at each recall $R$ by taking the maximum precision for any recall level $\tilde{R}\geq R$:
\mbox{$\hat P(R) = \max_{\tilde{R}\geq R}P(\tilde{R})
\label{eq:p_inter}
$},
where $P(\tilde{R})$ is the precision at recall $\tilde{R}$. The mean Average Precision (mAP) is computed over the number of unique classes. 
We define a true positive using the IoU, or Jaccard Index, between the predicted ($\tilde{b}$) and annotated 3D bounding box ($b$): \mbox{$ J(\tilde{b},b) = (\tilde{b} \cap b)/(\tilde{b} \cup b)
\label{eq:iou}
$},
which satisfies the condition \mbox{$J(\tilde{b},b) \geq \tau$} (e.g.~\mbox{$\tau = 50\%$}).
Note that we only consider the detection with the correct class and highest IoU as a true positive if there are multiple predictions above the threshold.

We evaluate \textit{object pose in 3D} by computing AP for predictions whose translation and rotation errors are less than a respective threshold~\cite{Wang2019CVPR_NOCS}. The \textit{translation error} is defined as the Euclidean distance between the predicted and annotated translation vectors, $\varepsilon_{t}= \|\mathbf{\tilde{t}} - \mathbf{t} \|_2 \in \mathbb{R}^3$. The \textit{rotation error} is given by
\mbox{$\varepsilon_{R} = \arccos{(( Tr(\mathbf{\tilde{R}} \mathbf{R}^T) - 1)/2)}$}, where $\tilde{\mathbf{R}} \in SO(3)$ ($\mathbf{R}$) is the predicted (annotated) $3\times3$ rotation matrix, $Tr(\cdot)$ is the trace operator, and $^T$ is the transpose operator. Note that we convert $\varepsilon_{\mathbf{R}}$ from radians to degrees for the discussion of the results.
We define a true positive as $\varepsilon_\mathbf{t} < \tau_\mathbf{t}$ and $\varepsilon_{\mathbf{R}} < \tau_{\mathbf{R}}$.

\begin{table}[t!]
\caption{Average Precision results of the re-trained NOCS model using Umeyama's algorithm for 3D object detection and 6D pose estimation per object category on the mixed-reality test set. 
}
\footnotesize
\centering
\begin{tabular}{l rr rrrr}
    \toprule
     & $J_{25}$ & $J_{50}$ & $\varepsilon_{5,5}$ & $\varepsilon_{10,5}$ & $\varepsilon_{10,10}$ & $\varepsilon_{15,10}$ \\
    \toprule
    \textit{box} & 99.3 & 92.9 & 2.9 & 24.7 & 24.7 & 53.3  \\
    \textit{nonstem} & 62.9 & 61.6 & 10.8 & 36.3 & 36.3 & 58.3 \\
    \textit{stem} & 85.8 & 84.3 & 9.7 & 34.1 & 34.2 & 53.3\\
    \midrule
    Mean & 82.7 & 79.6 & 7.8 & 31.7 & 31.7 & 55.0  \\
    \bottomrule
\end{tabular}
\label{tab:syntest-all}
\end{table}

\pgfplotstableread{som_iou.txt}\somiou
\pgfplotstableread{som_rot_translationfixed.txt}\somrot
\pgfplotstableread{som_tra_rotationfixed.txt}\somtra

\pgfplotstableread{ccm_iou.txt}\ccmiou
\pgfplotstableread{ccm_rot_translationfixed.txt}\ccmrot
\pgfplotstableread{ccm_tra_rotationfixed.txt}\ccmtra

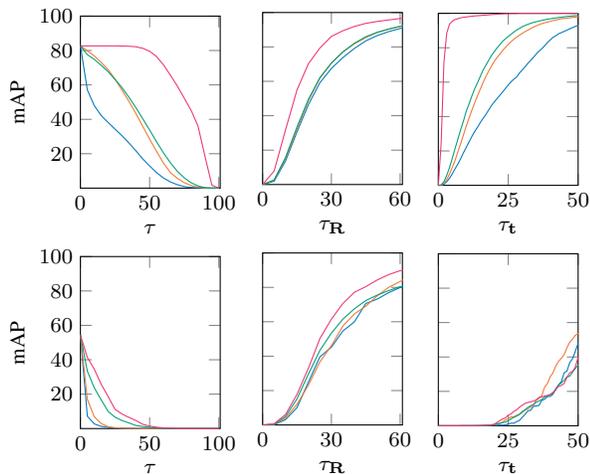
\begin{figure}[b!]
\footnotesize
\centering
\begin{tikzpicture}
    \begin{axis}[
        width=0.4\columnwidth,
        height=0.45\columnwidth,
        ymin=0,ymax=100,
        ytick={20,40,60,80,100},
        ylabel={mAP},
        tick label style={font=\footnotesize},
        xmin=0, xmax=101,
        xtick={0, 50, 100},
        xlabel={$\tau$},
    ]
    \addplot[color=c1] table[x=Threshold,y=EPnP-A] {\somiou};
    \addplot[color=c4] table[x=Threshold,y=EPnP-G] {\somiou};
    \addplot[color=c9] table[x=Threshold,y=EPnP-U] {\somiou};
    \addplot[color=c6] table[x=Threshold,y=Umey] {\somiou};
    \end{axis}
\end{tikzpicture}
\begin{tikzpicture}
    \begin{axis}[
        width=0.4\columnwidth,
        height=0.45\columnwidth,
        ymin=0,ymax=100,
        ytick={20,40,60,80,100},
        yticklabels={},
        tick label style={font=\footnotesize},
        xmin=0, xmax=61,
        xtick={0, 30, 60},
        xlabel={Rotation (degrees)},
        xlabel={$\tau_{\mathbf{R}}$}
    ]
    \addplot[color=c1] table[x=Threshold,y=EPnP-A] {\somrot};
    \addplot[color=c4] table[x=Threshold,y=EPnP-G] {\somrot};
    \addplot[color=c9] table[x=Threshold,y=EPnP-U] {\somrot};
    \addplot[color=c6] table[x=Threshold,y=Umey] {\somrot};
    \end{axis}
\end{tikzpicture}
\begin{tikzpicture}
    \begin{axis}[
        width=0.4\columnwidth,
        height=0.45\columnwidth,
        ymin=0,ymax=100,
        ytick={20,40,60,80,100},
        yticklabels={},
        tick label style={font=\footnotesize},
        xmin=0, xmax=50,
        xtick={0, 25, 50},
        xlabel={$\tau_{\mathbf{t}}$},
    ]
    \addplot[color=c1] table[x=Threshold,y=EPnP-A] {\somtra};
    \addplot[color=c4] table[x=Threshold,y=EPnP-G] {\somtra};
    \addplot[color=c9] table[x=Threshold,y=EPnP-U] {\somtra};
    \addplot[color=c6] table[x=Threshold,y=Umey] {\somtra};
    \end{axis}
\end{tikzpicture}

\begin{tikzpicture}
    \begin{axis}[
        width=0.4\columnwidth,
        height=0.45\columnwidth,
        ymin=0,ymax=100,
        ytick={20,40,60,80,100},
        ylabel={mAP},
        tick label style={font=\footnotesize},
        xmin=0, xmax=101,
        xtick={0, 50, 100},
        xlabel={$\tau$},
    ]
    \addplot[color=c1] table[x=Threshold,y=EPnP-A] {\ccmiou};
    \addplot[color=c4] table[x=Threshold,y=EPnP-G] {\ccmiou};
    \addplot[color=c9] table[x=Threshold,y=EPnP-U] {\ccmiou};
    \addplot[color=c6] table[x=Threshold,y=Umey] {\ccmiou};
    \end{axis}
\end{tikzpicture}
\begin{tikzpicture}
    \begin{axis}[
        width=0.4\columnwidth,
        height=0.45\columnwidth,
        ymin=0,ymax=100,
        ytick={20,40,60,80,100},
        yticklabels={},
        tick label style={font=\footnotesize},
        xmin=0, xmax=61,
        xtick={0, 30, 60},
        xlabel={$\tau_{\mathbf{R}}$},
    ]
    \addplot[color=c1] table[x=Threshold,y=EPnP-A] {\ccmrot};
    \addplot[color=c4] table[x=Threshold,y=EPnP-G] {\ccmrot};
    \addplot[color=c9] table[x=Threshold,y=EPnP-U] {\ccmrot};
    \addplot[color=c6] table[x=Threshold,y=Umey] {\ccmrot};
    \end{axis}
\end{tikzpicture}
\begin{tikzpicture}
    \begin{axis}[
        width=0.4\columnwidth,
        height=0.45\columnwidth,
        ymin=0,ymax=100,
        ytick={20,40,60,80,100},
        yticklabels={},
        tick label style={font=\footnotesize},
        xmin=0, xmax=50,
        xtick={0, 25, 50},
        xlabel={$\tau_{\mathbf{t}}$},
    ]
    \addplot[color=c1] table[x=Threshold,y=EPnP-A] {\ccmtra};
    \addplot[color=c4] table[x=Threshold,y=EPnP-G] {\ccmtra};
    \addplot[color=c9] table[x=Threshold,y=EPnP-U] {\ccmtra};
    \addplot[color=c6] table[x=Threshold,y=Umey] {\ccmtra};
    \end{axis}
\end{tikzpicture}
\caption{Mean Average Precision (mAP) results while varying the threshold for 3D IoU ($\tau$, \%), object orientation ($\tau_{\mathbf{R}}$, degrees) or translation ($\tau_{\mathbf{t}}$, cm) on the mixed-reality test set (top) and the CORSMAL Containers Manipulation dataset (bottom). Legend:
EPnP-A ({\protect\raisebox{2pt}{\protect\tikz \protect\draw[c1,line width=2] (0,0) -- (0.3,0);}}),
EPnP-G ({\protect\raisebox{2pt}{\protect\tikz \protect\draw[c4,line width=2] (0,0) -- (0.3,0);}}), 
EPnP-U ({\protect\raisebox{2pt}{\protect\tikz \protect\draw[c9,line width=2] (0,0) -- (0.3,0);}}), 
Umeyama ({\protect\raisebox{2pt}{\protect\tikz \protect\draw[c6,line width=2] (0,0) -- (0.3,0);}}).
}
\label{fig:thresholdanalysis}
\end{figure}


\subsection{Results and discussion}

We show the performance of the re-trained NOCS model with Umeyama for 3D object detection and 6D pose estimation on the test set of our mixed-reality dataset in Table~\ref{tab:syntest-all}. The model obtains a mAP of 55 percentage points (p.p.) across the three categories with an error under 15$^\circ$ and 10~cm. This performance decreases as we decrease the thresholds for orientation and translation. This shows that it is still challenging for such a model to achieve accurate pose estimations and correct object localisations, even on composite images of a mixed-reality dataset and despite the presence of clean and accurate depth maps. The category \textit{nonstem} is the most affected among the categories, especially due to false positives caused by inaccurate localisation or misclassification. However, performance is not significantly decreasing when increasing the threshold on the Jaccard index from 25 to 50. On the contrary, pose performance is more affected with decreases of about 20 p.p. in mAP for a more restrictive threshold every 5$^\circ$ (e.g., from 15 to 5). With the most restrictive threshold, $\varepsilon_{5,5}$, which would be an appropriate error range for potential human-robot collaboration, we can observe that the category \textit{box} becomes the least accurate, whereas the model retains 9.7\% and 10.8\% in mAP for \textit{stem} and \textit{nonstem}.


We compare the results of Umeyama, EPnP-A, EPnP-G, and EPnP-U, as we vary the thresholds on the mixed-reality test set and on the annotated CCM sequences in Fig.~\ref{fig:thresholdanalysis}.
Umeyama predicts more accurately poses and sizes than the EPnP variants thanks to the clean depth in the mixed-reality dataset, but this discrepancy reduces on CCM due to noisy depth images from the sensor and presence of transparent containers. We noticed that hand-occlusions may cause misclassification between  \textit{nonstem} and \textit{stem}. 
When generalising to real data and a new scene, it is hard for  the model coupled with the pose recovery strategies to preserve similar performance as in the mixed-reality test set. This is shown by a drop in both translation and IoU as the thresholds are more restrictive.

\begin{figure}[t!]
    \centering
    \includegraphics[width=0.9\columnwidth]{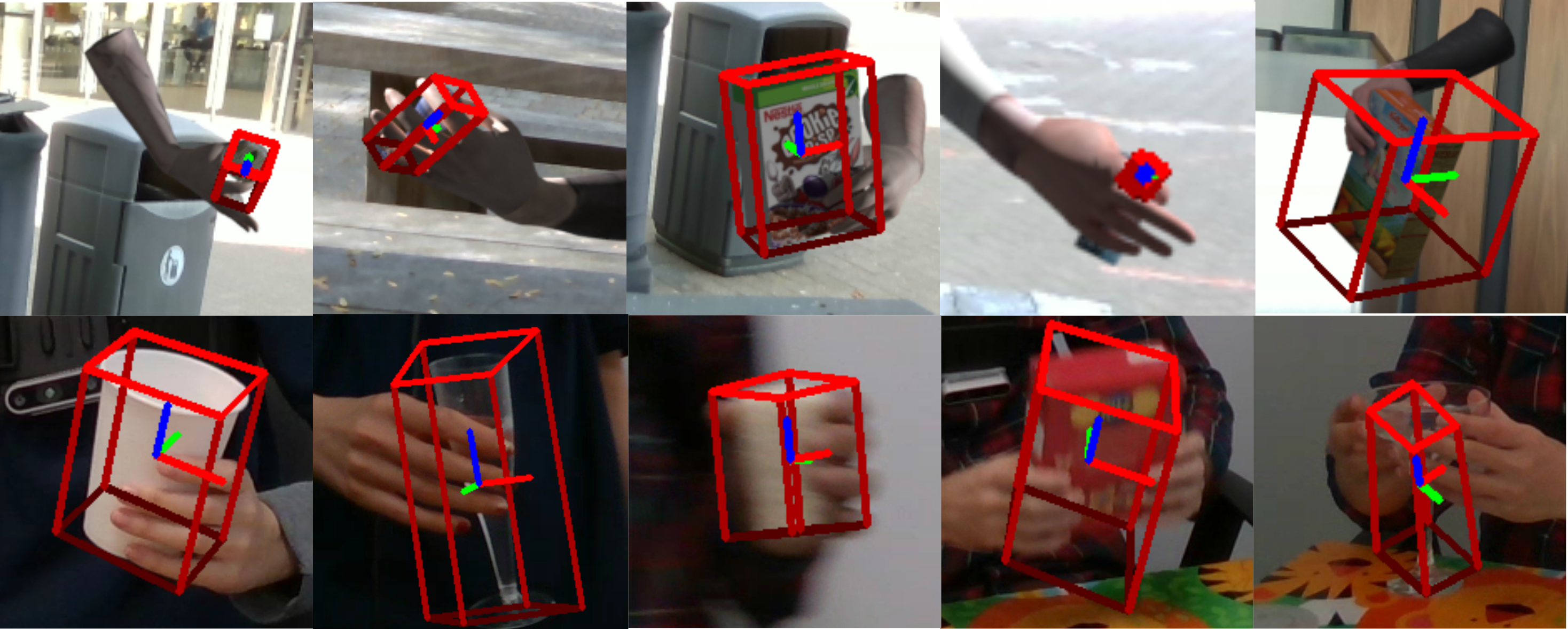}
    \caption{Sample results (image crops) of 6D pose estimation with NOCS and Umeyama on the mixed-reality test set (top) and with NOCS and EPnP-A on the real sequences of CCM (bottom).}
    \label{fig:visres_qual}
    \vspace{-10pt}
\end{figure}

Fig.~\ref{fig:visres_qual} shows sample results of 6D pose estimation on both testing datasets. Note some large inaccuracies in the last two columns due to heavy occlusions (small-size object), overestimated size, challenging object pose, or the gap between synthetic and real data.



\section{CONCLUSION}

We presented a new mixed-reality dataset and annotated sequences of the CORSMAL Containers Manipulation dataset~\cite{Xompero2022Access} for the task of category-level 6D pose estimation of hand-occluded containers. We showed that re-training a model on our dataset improves the  accuracy of poses and sizes of a set of object categories. However, generalisation to a real scene is still challenging due to noise in the depth data, transparencies and clutter.  We hope that the dataset and baseline model will facilitate research and benchmarking. 

Future work includes reducing the dependency on depth data  and the application of the model to human-robot collaboration.


\bibliographystyle{IEEEbib}
\bibliography{main}

\end{document}